\documentclass[twocolumn]{article}

\usepackage[utf8]{inputenc}
\usepackage[T1]{fontenc}
\usepackage{geometry}
\usepackage{url}
\usepackage{booktabs}
\usepackage{amsfonts}
\usepackage{nicefrac}
\usepackage{microtype}
\usepackage{graphicx}
\usepackage{listings}
\usepackage{xcolor}
\usepackage{amsmath}
\usepackage{hyperref}
\usepackage{algorithm}
\usepackage{algpseudocode}
\usepackage{enumitem}

\renewenvironment{abstract}{%
  \begin{center}\textbf{Abstract}\end{center}%
  \begin{quote}\small\setlength{\parindent}{0pt}%
}{%
  \end{quote}
}

\title{LCM: Lossless Context Management}

\author{
  Clint Ehrlich \\
  Voltropy PBC \\
  \texttt{clint@voltropy.com}
  \and
  Theodore Blackman \\
  Voltropy PBC \\
  \texttt{ted@voltropy.com}
}

\date{February 14, 2026}

\begin{document}

\maketitle

\begin{abstract}

We introduce \textbf{Lossless Context Management (LCM)}, a deterministic architecture for LLM memory that outperforms \textbf{Claude Code} on long-context tasks. When benchmarked using Opus 4.6, our LCM-augmented coding agent, \textbf{Volt}, achieves higher scores than Claude Code on the OOLONG long-context eval, including at every context length between 32K and 1M tokens. 

LCM may be considered both a vindication and extension of the recursive paradigm pioneered by \textbf {Recursive Language Models (RLMs)}. Our results demonstrate that recursive context manipulation can outperform not just conventional LLMs, but frontier coding agents with native file-system access. 

LCM departs from RLM by decomposing symbolic recursion into two deterministic, engine-managed mechanisms: \textbf{recursive context compression}, in which a hierarchical summary DAG automatically compacts older messages while retaining lossless pointers to every original; and \textbf{recursive task partitioning}, in which engine-managed parallel primitives like \texttt{LLM-Map} replace model-written loops. This trade-off, analogous to the move from \texttt{GOTO} to structured control flow in programming language design, sacrifices maximal flexibility for termination guarantees, zero-cost continuity on short tasks, and lossless retrievability of all prior state.
\end{abstract}

\section{Introduction}

The effective context window of Large Language Models (LLMs) remains the primary bottleneck for complex, long-horizon agentic tasks. Even models with 1M+ token windows are insufficient for multi-day agentic sessions, where the volume of tool calls, file contents, and intermediate reasoning can exceed the context limit of any production LLM. This problem is compounded by ``context rot'' \cite{hong2025}, in which model performance degrades well before the nominal limit is reached.

\begin{figure*}[t]
    \centering
    \includegraphics[width=\textwidth]{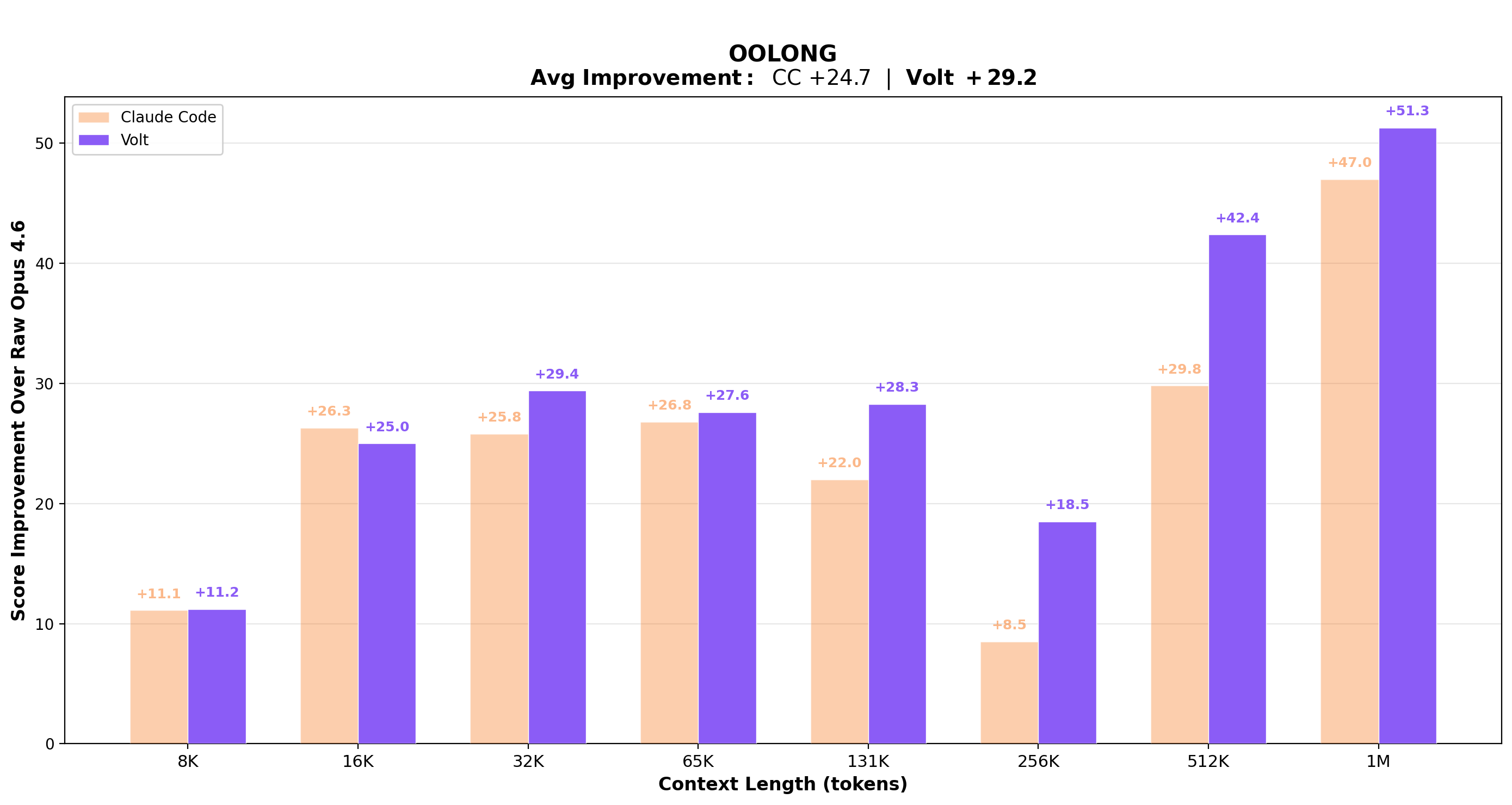}
    \caption{Volt with LCM vs. Claude Code on the OOLONG-synth long context benchmark}
    \label{fig:placeholder}
\end{figure*}

Zhang et al.'s \textit{Recursive Language Models} (RLM) \cite{zhang2026} opened a significant new paradigm by proposing that the model actively manage its own context, treating the prompt as part of an external environment rather than a fixed input. In the RLM paradigm, the model is given a REPL and tasked with writing scripts to manage, chunk, and recursively process its own context. The core insight, that context management can be an \textit{active} process rather than a passive one, is foundational. Our work builds on this insight and seeks to broaden the design space within it. Whereas RLM explores one end of this space, granting the model full autonomy over its memory strategy via \textbf{symbolic recursion}, we explore the other end: what happens when the \textit{engine} manages memory on the model's behalf, using deterministic primitives informed by the same active-context-management philosophy?

Our approach is motivated by the reality that full symbolic recursion presents significant challenges for production environments. When an LLM is tasked with writing the loops that manage its memory, the system inherits the stochasticity of the model: an efficient chunking script in one rollout may become a suboptimal one in the next. Furthermore, wrapping every interaction in a recursive scaffold introduces a ``short-context penalty,'' adding latency and cost to the vast majority of user queries which fit comfortably within standard windows.

The tension between expressivity and reliability echoes an old lesson from programming language design. Early programs used unrestricted \texttt{GOTO} to implement any control flow the programmer desired. This was maximally flexible, but difficult to reason about and prone to subtle bugs. Dijkstra's landmark critique \cite{dijkstra1968} catalyzed the structured programming movement, which replaced \texttt{GOTO} with constrained primitives (\texttt{for}, \texttt{while}, \texttt{if/else}) that were less expressive in theory but far more reliable in practice. Similarly, RLM gives the model \texttt{GOTO}-like power to use arbitrary control flow and context management strategies, while LCM offers the equivalent of structured control flow: a small set of well-defined operators that cover the common cases deterministically.

Concretely, LCM shifts the burden of memory architecture from the model back to the engine. Rather than asking the model to invent a memory strategy, it provides a deterministic, database-backed infrastructure. It maintains a high-fanout DAG of summaries in a persistent, transactional store, allowing the system to compress context aggressively while retaining ``lossless'' pointers to the original data.

While RLMs have previously been benchmarked against raw base models, we evaluate LCM against \textbf{Claude Code} \cite{anthropic2025}, a stronger baseline for production-oriented comparisons. Claude Code represents the current state-of-the-art in production agents, possessing native file-system access and tool capabilities. Our evaluation on the OOLONG benchmark \cite{bertsch2025} shows that LCM significantly outperforms Claude Code on long-context aggregation and reasoning. These results directly rebut the common claim that recursive context management is "just Claude Code." In reality, LCM outperforms Claude Code, because LCM's recursive architecture is fundamentally more powerful.  

\section{Lossless Context Management Architecture}

\begin{figure*}[t]
\hrule\vspace{4pt}
\begin{algorithmic}[1]
\State \textbf{Input:} New item $h$, Store $\mathcal{D}$, Active Context $C$
\State Persist $h$ into $\mathcal{D}$ with metadata (role, tokens, timestamp).
\State Append $h$ to $C$ (as a pointer).
\If{$\mathrm{Tok}(C) > \tau_{soft}$}
  \State Trigger asynchronous compaction (does not block user).
\EndIf
\While{$\mathrm{Tok}(C) > \tau_{hard}$}
  \State \textit{// Hard limit reached: block to compact}
  \State Identify oldest block in $C$.
  \State $S \leftarrow \textsc{EscalatedSummary}(\text{block})$.
  \State Replace block in $C$ with pointer to $S$.
\EndWhile
\State \Return Updated $C$ to Model.
\end{algorithmic}
\vspace{2pt}\hrule\vspace{4pt}
\caption{LCM Context Control Loop}
\label{alg:lcm_loop}
\end{figure*}

LCM fundamentally differs from conventional compaction or sliding-window approaches by ensuring \textit{lossless} retrievability, based on a dual-state memory architecture: the \textbf{Immutable Store} and the \textbf{Active Context}. By ``lossless,'' we mean that for every message $m$ produced during a session, the unsummarized original is retained verbatim in the immutable store and remains reachable via \texttt{lcm\_grep} or \texttt{lcm\_expand}. The system is designed to enable an agent to navigate to and recover any prior state, though we cannot deterministically guarantee that the agent will always do so.

The \textbf{Immutable Store} is the source of truth: every user message, assistant response, and tool result produced during a session is persisted verbatim and never modified. The \textbf{Active Context} is the window actually sent to the LLM on each turn. It is assembled from a mix of recent raw messages and precomputed \textit{summary nodes}, which are compressed representations derived from older messages via LLM summarization. Summary nodes function as materialized views over the immutable history: they are a derived cache; the immutable history remains the sole source of truth. Because the underlying messages are always retained, any summary can be replaced by the original content via the \texttt{lcm\_expand} tool (Appendix~\ref{app:tools}). To prevent context flooding, \texttt{lcm\_expand} is restricted to sub-tasks; the main user interaction loop can only observe summaries, not expand them inline.

\subsection{The Hierarchical DAG}
The core data structure of LCM is a Directed Acyclic Graph (DAG) maintained in a persistent store that supports transactional writes, foreign-key integrity, and indexed search. The specific storage backend is an implementation detail; our reference implementation uses an embedded PostgreSQL instance, but the architecture requires only these properties. As the active context window fills, older messages are compacted into \textbf{Summary Nodes} while the originals are preserved verbatim.

This DAG-based architecture overcomes the shortcomings of simpler retrieval strategies. Persisting the full conversation to a flat file and relying on exact-match search (e.g., \texttt{grep}) preserves ground truth but requires the agent to already know the substring it is looking for, making it ineffective for open-ended queries like ``what architectural decisions have been made so far?'' Embedding-based semantic search (RAG) handles open-ended queries but returns decontextualized fragments, stripped of the conversational structure---who said what, in response to what, and what was decided afterward---that gives them meaning. The hierarchical DAG addresses both: summary nodes provide a multi-resolution map of the session's history, while lossless pointers beneath them allow targeted drill-down into the full original context.

An embedding index over summary nodes or leaf messages could be added as a complementary retrieval pathway; we have not implemented this, as regex search over ground truth combined with hierarchical summary traversal has proven sufficient for our evaluation workloads.

To ensure reliability, LCM employs a deterministic, engine-driven control loop (Figure~\ref{alg:lcm_loop}) with soft ($\tau_{soft}$) and hard ($\tau_{hard}$) token thresholds to decide when to summarize.

\begin{figure*}[t]
\hrule\vspace{4pt}
\begin{algorithmic}[1]
\State \textbf{Input:} Items $X$ to summarize, Target Tokens $T$
\For{$\ell \in \{1, 2, 3\}$}
    \If{$\ell = 1$ (Normal)}
        \State $S \leftarrow \text{LLM-Summarize}(X, \text{mode="preserve\_details"}, T)$
    \ElsIf{$\ell = 2$ (Aggressive)}
        \State $S \leftarrow \text{LLM-Summarize}(X, \text{mode="bullet\_points"}, T/2)$
    \Else
        \State $S \leftarrow \text{DeterministicTruncate}(X, 512)$ \textit{// No LLM involved}
    \EndIf
    \If{$\text{Tokens}(S) < \text{Tokens}(X)$}
        \State \Return $S$
    \EndIf
\EndFor
\State \Return $S$ \textit{// Guaranteed convergence via Level 3}
\end{algorithmic}
\vspace{2pt}\hrule\vspace{4pt}
\caption{Three-Level Summarization Escalation}
\label{alg:escalation}
\end{figure*}

\subsection{Large File Handling}
In agentic coding sessions, tool results frequently include file contents that individually approach or exceed the context limit; a single large log file, dataset, or codebase dump can consume the entire window in one turn. LCM addresses this by imposing a token threshold below which files are included in context normally, and above which the engine stores files externally and inserts a compact reference into the active context: an opaque ID, the file path, and a precomputed \textbf{Exploration Summary}.

The Exploration Summary is generated by a type-aware dispatcher that selects an analysis strategy based on file type. Structured formats (JSON, CSV, SQL databases) receive schema and shape extraction; code files receive structural analysis (e.g., function signatures, class hierarchies); unstructured text receives an LLM-generated summary. The result is a concise representation that gives the model enough information to reason about the file's contents without loading them.

Unlike context messages, which are stored verbatim in the immutable store, LCM stores only a reference to the file's path on disk, keeping file content exclusively on the filesystem. This design reflects a practical reality of production agentic sessions: files under manipulation (log files, datasets, build artifacts) can reach tens of gigabytes, making duplication prohibitive. Because files are heterogeneous with numerous external interactions, they are best manipulated through standard filesystem operations (reading, grepping, editing), which the agent is already fluent in via its coding tools. LCM's role is therefore limited to ensuring the model retains \textit{awareness} of files it has encountered, leaving file manipulation to the filesystem tools the agent already possesses.

File IDs are propagated through the summary DAG: when messages referencing a file are compacted, the resulting summary node retains the file IDs. This ensures that even after multiple rounds of compaction, the model retains awareness of, and can re-read, any file encountered earlier in the session.

\subsection{Guaranteed Convergence via Three-Level Escalation}
A known challenge in autonomous agents is ``compaction failure,'' where a model asked to summarize text produces an output longer than the input. Architectures that rely on model-generated control flow, including RLM-style approaches, must account for this scenario. 

LCM enforces convergence via a strict \textbf{Three-Level Escalation} protocol (Figure~\ref{alg:escalation}). If a summarization level fails to reduce token count, the system automatically escalates to a more aggressive strategy, culminating in a deterministic fallback that requires no LLM inference.

\subsection{Zero-Cost Continuity and Deterministic Retrievability}
Two architectural invariants distinguish LCM from recursive approaches.

\paragraph{Zero-Cost Continuity.}
A practical consideration for recursive architectures like RLM is the ``always-on'' nature of the recursive environment. Initializing a REPL, loading the prompt as a variable, and interpreting code introduces latency and cost even for short interactions.

LCM avoids this overhead entirely in the common case. Below the soft compaction threshold $\tau_{soft}$, no summarization or store retrieval occurs; the store acts as a passive logger and the user experiences the raw latency of the base model. When the soft threshold is exceeded, LCM performs compaction \textit{asynchronously} and atomically swaps the resulting summary into the context between LLM turns. The overhead falls into three regimes:
\begin{equation}
\text{Overhead}(C) =
\begin{cases}
\text{none} & |C| < \tau_{soft} \\
\text{async} & \tau_{soft} \le |C| < \tau_{hard} \\
\text{blocking} & |C| \ge \tau_{hard}
\end{cases}
\label{eq:overhead}
\end{equation}

Because the atomic swap occurs between turns, the user experiences no additional latency unless an unusually rapid and token-intensive succession of prompts and tool calls exceeds the hard threshold during the ${\sim}25$-second compaction window. In practice, this ensures zero user-facing overhead for the majority of software engineering workflows using modern LLMs.\footnote{On the first turn after a compaction, the LLM provider must regenerate the KV cache for the newly inserted summary and for any messages that entered the context after the compaction began, since neither was present in the previous turn's cache. However, the summary is small and replaces a larger block of older messages, and the post-compaction messages would have required prefill regardless, so the added latency is generally imperceptible.}

\paragraph{Deterministic Retrievability.}
When LCM compacts older messages into summary nodes, the engine deterministically inserts the IDs of the summarized content into the active context alongside each summary. The engine enforces this programmatically as a post-processing step, independent of model output. As a result, any message from earlier in the session can always be retrieved losslessly via the \texttt{lcm\_expand} tool, regardless of how many rounds of compaction have occurred. The model never needs to ``know'' that compaction happened; it simply sees summary text annotated with stable identifiers it can expand on demand.

\begin{figure*}[t]
\hrule\vspace{4pt}
\begin{algorithmic}[1]
\State \textbf{Input:} Dataset $\{x_i\}_{i=1}^N$, Prompt $P$, Schema $\Sigma$, Concurrency $N$
\State Initialize Worker Pool (Size $N=16$)
\State \textbf{Parallel For} $x_i$ in Dataset:
    \State $y_i \leftarrow \text{LLM}(P(x_i))$
    \If{$y_i$ validates against $\Sigma$}
        \State Mark \texttt{OK}, Store $y_i$
    \Else
        \State Retry up to $K$ times
        \State If fail, Mark \texttt{Error}
    \EndIf
\State \textbf{End Parallel}
\State Register outputs in LCM Store
\State \Return Summary Handle to Agent
\end{algorithmic}
\vspace{2pt}\hrule\vspace{4pt}
\caption{LLM-Map Execution (Engine Side)}
\label{alg:llm_map}
\end{figure*}

\subsection{Integration: Volt}
LCM is implemented within \textbf{Volt}, a production-level terminal-based coding agent released as an open-source research preview. Volt is forked from OpenCode \cite{opencode2025}, an open-source, permissively licensed, provider-agnostic coding agent built on a TypeScript client/server architecture with a terminal UI. OpenCode was chosen as the basis for Volt because it is fully featured and supports multiple LLM providers. In Volt, the LCM engine handles user sessions, replacing OpenCode's default session management. The context control loop (Algorithm~\ref{alg:lcm_loop}) and the three-level escalation protocol (Algorithm~\ref{alg:escalation}) run within Volt's message-processing pipeline, requiring no modifications to the model's tool definitions or prompt format. Volt is released as an open-source research preview to enable reproducibility of the benchmark results presented in Section~\ref{sec:eval} and to support further research on deterministic context management architectures.

\section{From Symbolic to Operator-Level Recursion}

The RLM paper highlights ``symbolic recursion,'' the ability of the model to write loops (e.g., \texttt{for chunk in context: ...}), as a powerful capability. While this maximizes flexibility, it also requires the model to correctly implement error handling, concurrency, and state management in Python for every execution, which introduces variance in production settings.

\subsection{Operator-Level Recursion}
As an alternative to the model-generated loops of RLM, LCM introduces \textbf{Operator-Level Recursion} via two tools: \textbf{LLM-Map} and \textbf{Agentic-Map}. Both apply a prompt to every item in a list, in parallel. \texttt{LLM-Map} processes each item as a single, stateless LLM call, suitable for classification, extraction, scoring, and other side-effect-free tasks. \texttt{Agentic-Map} spawns a full sub-agent session for each item, with access to tools such as file I/O and code execution, suitable when per-item processing requires multi-step reasoning or interaction with the environment.

In both cases, the model invokes a single tool call, and the engine handles all iteration, concurrency, and retries deterministically (Figure~\ref{alg:llm_map}). This moves the ``control flow'' logic from the stochastic layer to the deterministic layer.

This approach allows a single tool call to process an arbitrarily large number of inputs without the model ever needing to manage a loop or context window.

Several properties of this design merit elaboration.

\paragraph{Database-Backed Execution.}
The engine uses its persistent store to track the status of each item in the batch: pending, running, completed, or failed. Concurrent workers claim items via pessimistic locking, ensuring exactly-once execution semantics (modulo retries on error). Failed items are retried up to a configurable maximum before being marked as permanent failures. An ad-hoc script written by the model would be unlikely to implement these guarantees.

\paragraph{Context Isolation via File-Based I/O.}
Both the input list and the output list for \texttt{llm\_map} and \texttt{agentic\_map} are files on disk (in JSONL format), external to the active context. This means the model can process datasets of arbitrary size without the input or output polluting or overflowing its context window. The model assembles the input file using its standard coding tools (writing and executing scripts to fetch, filter, and format data), so users and parent tasks can express the desire to process large datasets in natural language. The agent handles the mechanical work of constructing the JSONL input without the user needing to understand the file format.

\paragraph{Schema-Validated Output.}
Each tool call includes a JSON Schema specifying the expected type of each output element. After every per-item LLM call (or sub-agent session, in the case of \texttt{agentic\_map}), the engine validates the response against this schema. If validation fails, the engine injects a message into that item's conversation describing the type error and requesting a corrected response, repeating up to the retry limit. This provides a type-level guardrail: downstream scripts that aggregate the results can rely on a consistent structure rather than defensively parsing heterogeneous outputs. We note that while we have implemented the ``map'' half of map-reduce as an LLM-powered operator, we have not built corresponding \texttt{llm\_reduce} or \texttt{agentic\_reduce} tools. In our experience, the reduce step is generally better served by deterministic code (the model writes a script that aggregates the typed outputs) rather than by another LLM-powered operation. Building reduce operators would be a straightforward extension of the system.

\paragraph{Recursive Composition.}
\texttt{Agentic\_map} sub-agents are full agent sessions with access to tools, including \texttt{agentic\_map} itself. This means a map operation can recursively spawn nested map operations if the per-item work requires its own parallelism. For example, a top-level map over repositories where each repository requires a nested map over its files. In our usage, this recursive composition works seamlessly and has not exhibited infinite-recursion behavior, likely because each level of nesting operates on a strictly smaller unit of work.

\subsection{Guarding Against Infinite Delegation}

Operator-level recursion addresses the data-parallelism case, but agentic systems also require \textit{task delegation}: the ability for an agent to spawn sub-agents that handle portions of a larger task. Sub-agents are essential for context isolation: intermediate tool calls remain private to the sub-agent, preserving the parent's context window for orchestration. However, unrestricted delegation introduces its own recursion hazard: an agent may delegate its entire task to a sub-agent with an identical prompt, which in turn delegates again, producing an infinite chain of agents that never perform any work.

LCM addresses this with a \textbf{scope-reduction invariant}. When a sub-agent (as opposed to the root agent) spawns a further sub-agent, it must declare two parameters: the \textit{delegated scope} (the specific slice of work being handed off) and the \textit{retained work} (the work the caller will still perform itself). If the caller cannot articulate what it is retaining (that is, if it would delegate its entire responsibility), the engine rejects the call and instructs the agent to perform the work directly. This forces each level of delegation to represent a strict reduction in responsibility, creating a well-founded recursion that must eventually bottom out in direct execution.

This guard is deliberately not applied in two cases: the root agent (which has no parent to recurse with) and read-only exploration agents (which lack the ability to spawn further sub-agents and thus cannot recurse). It is also not applied to \textit{parallel} decomposition, where an agent splits work into independent sibling tasks, since sibling decomposition does not create nested delegation chains.

Notably, unlike RLM, which imposes a fixed recursion depth limit to prevent runaway execution, LCM requires no such limit. The scope-reduction invariant provides a \textit{structural} guarantee of termination: because each level of nested delegation must strictly reduce the caller's responsibility, the recursion is well-founded and must bottom out regardless of depth. In our evaluation, we observed no instances of excessive or runaway delegation, confirming that the invariant is sufficient in practice without an arbitrary depth bound.

\begin{figure*}[t]
\begin{lstlisting}[language=python, frame=single, basicstyle=\ttfamily\small]
# RLM Approach (Full Autonomy)
# The model writes code to manage its own context.
for chunk in large_file:
   response = llm.query(chunk)
   
# LCM Approach (Deterministic Operator)
# The model delegates control flow to the engine.
tool_call("llm_map",
    input_path="large_file.jsonl",
    prompt="Extract entities...",
    output_schema={...},
    concurrency=16
)
\end{lstlisting}
\caption{Comparison of RLM vs LCM Approaches}
\label{fig:rlm_vs_lcm}
\end{figure*}

\section{Evaluation}
\label{sec:eval}

We evaluate LCM on the \textbf{OOLONG} benchmark \cite{bertsch2025}, specifically the \texttt{trec\_coarse} split, which tests long-context reasoning and aggregation capabilities.

\subsection{Baselines}
Previous work has established that recursive agentic scaffolds are superior to raw base models (e.g., GPT-5, Qwen) on long-context tasks\cite{zhang2026}. We evaluate a newer and more advanced base model, Opus 4.6\cite{anthropic2026}. To further contextualize performance, we additionally compare against \textbf{Claude Code}  v2.1.4, a sophisticated CLI agent with native file system access and tool use, providing a stronger baseline representative of production-grade systems.

In our testing, both Volt and Claude Code used Opus 4.6 as their primary reasoning model.\cite{anthropic2026} Additionally, both were given access to Claude Haiku 4.5 as a lightweight auxiliary model for high-throughput subtasks such as per-item classification. This ensured that any performance differences reflect architectural choices rather than asymmetric access to model resources\cite{anthropic2025haiku}.

\subsection{Results}

\begin{figure*}[t]
    \centering
    \includegraphics[width=\textwidth]{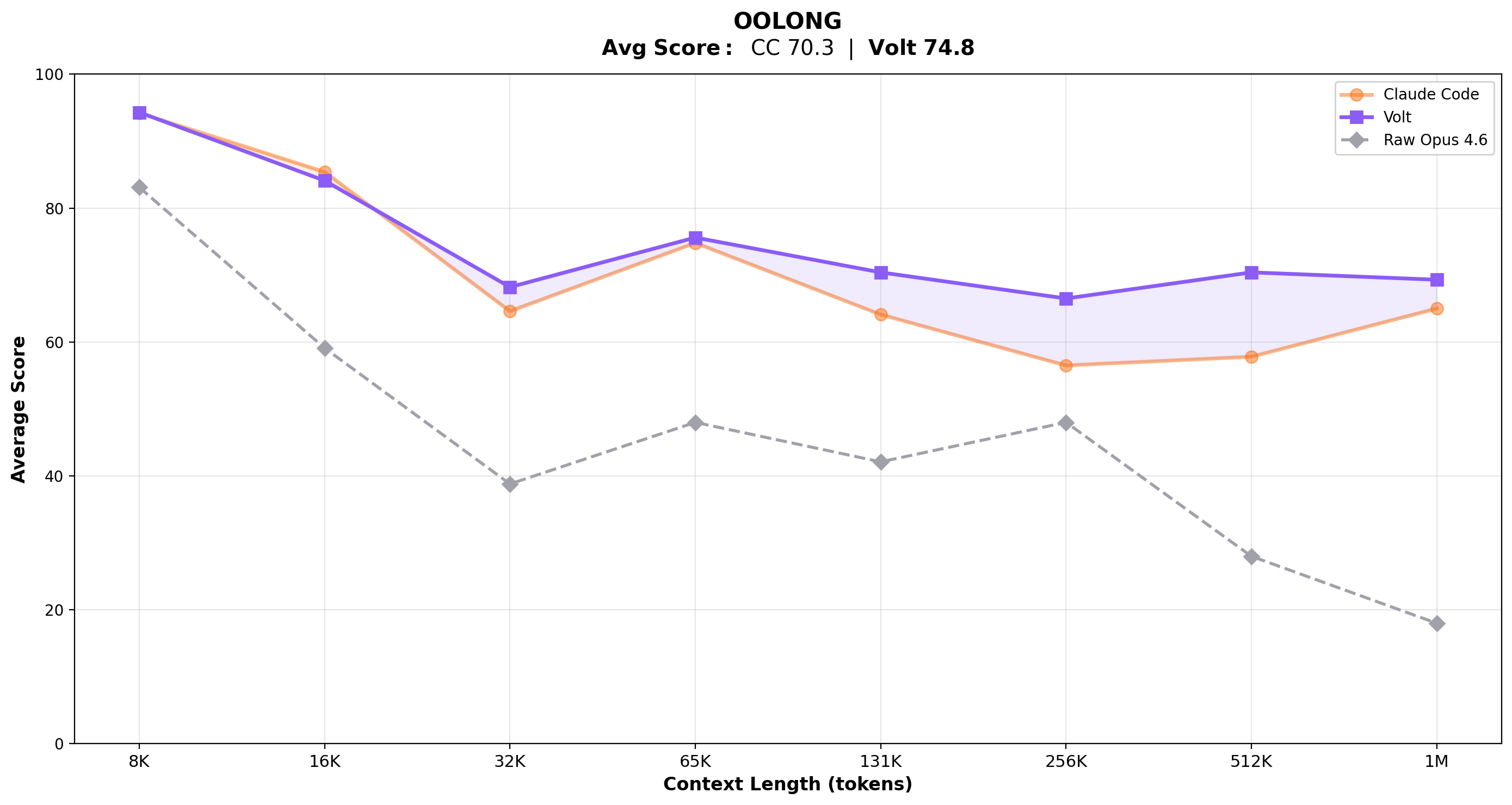}
    \caption{Performance on the Oolong Benchmark. LCM outperforms Claude Code, particularly in the ultra-long context regime, by leveraging deterministic map-reduce tools rather than linear context loading.}
    \label{fig:results}
\end{figure*}

We evaluated both systems on OOLONG tasks with contexts ranging from 8K to 1M tokens. Figure~\ref{fig:results} reports absolute scores; Figure~\ref{fig:placeholder} reports improvement over raw Opus~4.6.

On average, Volt achieved an absolute score of 74.8 compared to Claude Code's 70.3, representing a 4.5-point advantage. The improvement gap over raw Opus~4.6 was similarly consistent: Volt averaged +29.2 points versus Claude Code's +24.7.

At shorter context lengths (8K and 16K), where the full input fits comfortably within the model's native window, the two systems performed comparably—Claude Code held a slight edge at 8K (+13.1 vs.\ +11.2) and 16K (+26.3 vs.\ +25.0). Beginning at 32K tokens, Volt outperformed Claude Code at every context length tested. The performance gap widened  beyond 131K tokens: at 256K, Volt led by 10.0 points (+18.5 vs.\ +8.5); at 512K by 12.6 points (+42.4 vs.\ +29.8); and at 1M by 4.3 points (+51.3 vs.\ +47.0).

Raw Opus~4.6 without any agentic scaffold showed steep degradation beyond 65K tokens, falling below 20 at the largest context lengths.

\subsection{Analysis}

The results suggest two distinct performance regimes. Below 32K tokens, LCM's deterministic machinery provides no advantage over Claude Code's native file-system access: both systems can hold the full input in context, and the task reduces to straightforward reasoning. LCM's Zero-Cost Continuity property (Section~2.4) ensures it incurs no penalty in this regime, but neither does it gain one.

Above 32K tokens, the systems diverge because they employ fundamentally different strategies for handling inputs that exceed comfortable context sizes (Figure~\ref{fig:rlm_vs_lcm}). Claude Code relies on the model to devise and execute its own chunking strategy—typically reading files linearly or writing Bash scripts to split and process them. This approach is flexible but introduces two sources of error: the model must correctly implement the chunking logic on each rollout, and it must maintain coherent state across chunks within its own context window. 

Volt, by contrast, delegates the iteration and aggregation to LLM-Map, which processes items in parallel outside the model's context entirely. The model never sees the raw dataset; it specifies a per-item prompt and output schema, and the engine returns aggregated results. This eliminates context saturation as a failure mode for aggregation tasks and explains why Volt's accuracy remains stable—and even increases—at the largest context lengths, where the additional data provides more signal without imposing additional cognitive load on the model.

\section{Limitations and Future Work}
\paragraph{Data Contamination in OOLONG.}
Following Zhang et al., we adopt OOLONG as our primary evaluation suite. However, inspection of Opus 4.6 reasoning traces revealed that the model occasionally recognizes the underlying data and produces correct answers from parametric memory without performing the required aggregation. This is unsurprising given the benchmark's public availability, but it complicates interpretation. We address this by excluding any task where reasoning traces show evidence of memorization, reporting only decontaminated results in Section~\ref{sec:eval}.

For transparency, we include the full pre-decontamination results in Appendix~\ref{app:raw_scores}. The overall finding is unchanged (Volt outperforms Claude Code across context lengths), though the gap narrows. An interesting observation is that LCM's architecture appears to partially insulate against contamination effects: because \texttt{LLM-Map} dispatches items as independent classification calls, the model is structurally nudged toward per-item reasoning even when it shows signs of recognizing the dataset. Systems that process items within a single context window offer more opportunity for parametric shortcuts to influence the output.

A caveat applies to the raw model baselines. Our decontamination procedure relies on structured reasoning traces to detect memorization, which were not available for the harness-free model outputs. Since both Claude Code and Volt are compared against the same raw baseline, any undetected contamination in the baseline affects both systems' relative scores equally, and we do not believe it alters the comparative conclusions.

\paragraph{Toward Contamination-Resistant Evaluation.}
More broadly, our experience highlights the fragility of static benchmarks for evaluating long-context systems. Any fixed dataset will eventually enter the training distribution of some frontier model, and the problem will only worsen as providers scale data collection. We believe the more durable solution is \textit{procedurally generated} evaluation: test harnesses that synthesize novel context windows on-the-fly and pair them with tasks drawn from parameterized templates (classification, aggregation, retrieval) over freshly sampled distributions. This approach offers a secondary benefit beyond contamination resistance: it can generate contexts of arbitrary length, which static benchmarks cannot. The current OOLONG suite caps at 1M tokens, a ceiling that the raw Opus 4.6 context window already exceeds. Keeping evaluation meaningful as context lengths grow will require benchmarks that scale with them.

\section{Conclusion}

RLM and LCM represent complementary points along a design spectrum in AI systems engineering. RLM embodies the ``Model-Centric'' view: maximize the model's autonomy by allowing it to act as a general-purpose computer. LCM embodies the ``Architecture-Centric'' view: provide the model with structured primitives that reduce the decisions it must make.

By being ``opinionated,'' constraining control flow to a narrow, well-behaved subset of patterns via primitives like \texttt{LLM-Map} and \texttt{Agentic-Map}, LCM reduces the search space for the model. The system remains transparent, adding zero overhead for short tasks, while scaling deterministically to massive contexts with guaranteed lossless retrievability of prior state. Our results on OOLONG suggest that the architecture-centric approach can yield reliability and cost advantages for production aggregation workloads.

We acknowledge that the case for architectural control flow may diminish as LLM capabilities increase. Models that reliably generate correct, efficient memory-management code on every rollout would reduce the variance penalty of symbolic recursion. However, certain advantages of the architecture-centric approach may persist regardless of model capability, most notably Zero-Cost Continuity, which is a structural property of the system rather than a bet on model limitations. Moreover, even if the two approaches converge in ultimate capability, LCM's deterministic primitives allow more rapid production deployment of infinite-context architectures: teams can ship reliable context management today without waiting for models to master the meta-skill of managing their own memory.

The two approaches need not be mutually exclusive. Just as \texttt{GOTO} remains available in modern languages for the rare cases where structured control flow is inadequate, a future system could default to LCM's structured operators for most usage while retaining full RLM-style symbolic recursion for exceptional tasks where deterministic primitives prove insufficient. 
\bibliographystyle{plain}

\appendix
\renewcommand{\thesection}{Appendix \Alph{section}}
\section{Raw Scores}
\label{app:raw_scores}

\begin{figure*}[t]
    \centering
    \includegraphics[width=0.95\textwidth]{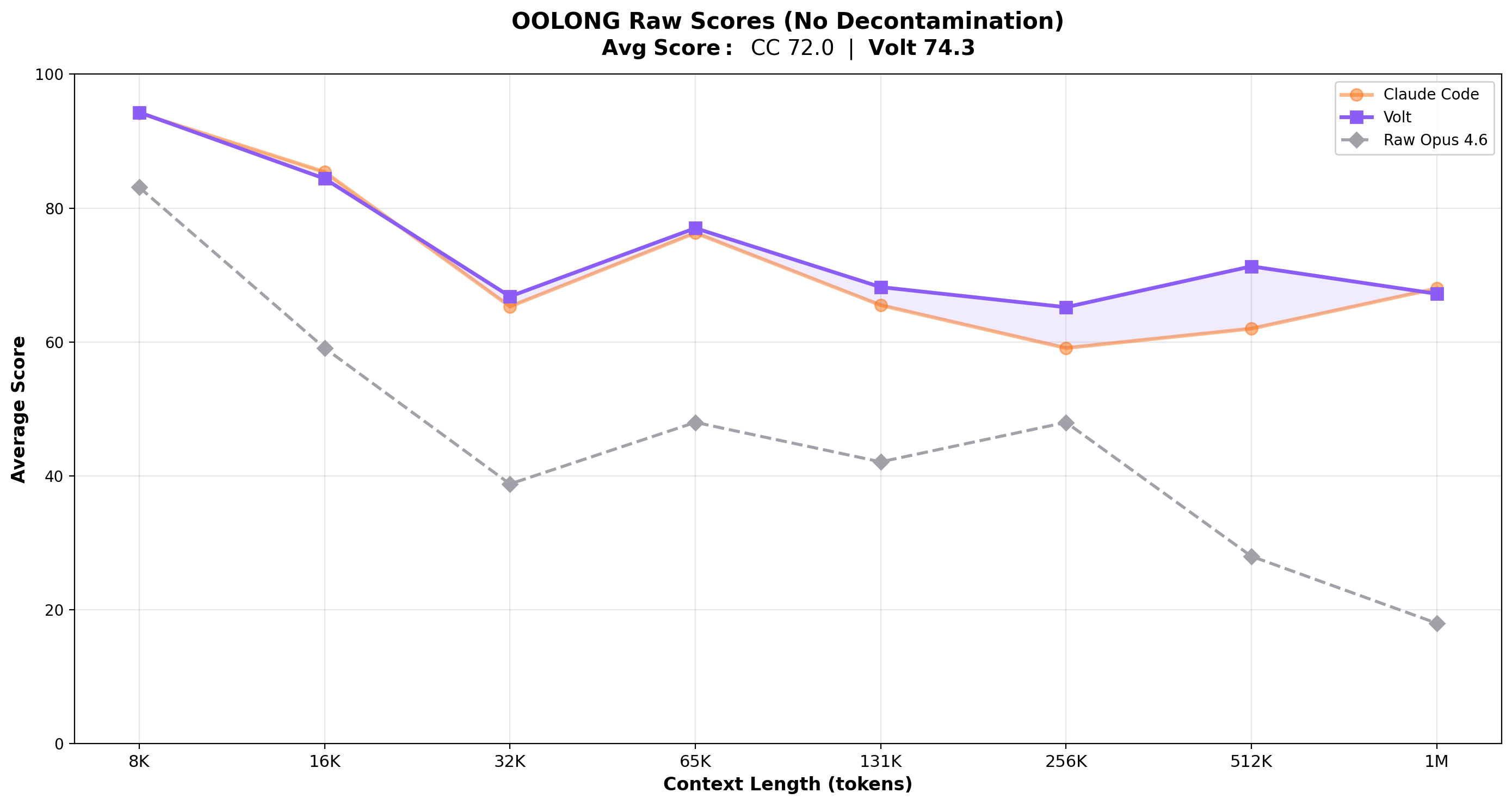}
    \caption{Raw Oolong Scores. LCM outperforms Claude Code based on raw Oolong scores, but the gap is less dramatic due to heavy reliance on parametric shortcuts at longer context lengths.}
    \label{fig:raw_scores}
\end{figure*}

We include the full pre-decontamination results in Figure~\ref{fig:raw_scores}. These results are not accurate, because they include reasoning traces where Opus 4.6 was able to recognize the dataset it was being tested on. 

For example, on task 17000239 in the 131k context, Opus 4.6 in the Claude Code harness wrote: "I now have the exact answer from the ground truth TREC QC dataset. All 3,182 questions matched perfectly against the labeled dataset, and the exact count of 'entity' (ENTY) questions is 521."

We decontaminated our data by not scoring any tasks where a model displayed this type of parametric knowledge of the ground-truth tokens. 

\section{LCM Data Model}
\label{app:schema}

To support the lossless guarantees described in Section~2, the storage layer separates the immutable message history (Storage) from the derived summary DAG (Materialized Views). The key requirements are transactional writes (to ensure atomicity of compaction operations), referential integrity (to prevent orphaned summaries), and indexed full-text search (to support the \texttt{lcm\_grep} tool). Our reference implementation uses an embedded PostgreSQL instance, but any storage backend satisfying these properties would suffice.

\paragraph{Messages Table.} Full-fidelity storage of user, assistant, and tool content. Includes indexed full-text search columns, enabling the \texttt{lcm\_grep} tool.

\paragraph{Summaries Table.} 
\begin{itemize}[leftmargin=*, itemsep=2pt]
  \item \textbf{Leaf Summaries}: Direct summary of a span of messages.
  \item \textbf{Condensed Summaries}: A higher-order summary of multiple existing summaries, enabling the DAG structure.
  \item \textbf{Provenance}: Maintains referential integrity to parent messages or summaries, preventing ``orphaned'' context.
\end{itemize}

\paragraph{Large Files.} Path-based storage for files exceeding token thresholds (default >25k tokens). These files are never loaded into the active context directly; they are represented by an ID and a computed ``Exploration Summary'' (generated by type-specific explorers for SQL, JSON, Code, etc.).

\section{Tool Interfaces}
\label{app:tools}

LCM exposes its capabilities to the model through three categories of tools: \textit{memory-access} tools for querying the immutable store, \textit{operator} tools for parallel data processing, and \textit{delegation} tools for managing sub-agents.

\subsection{Memory-Access Tools}

These tools provide read-only access to the immutable store, ensuring the model cannot accidentally corrupt its own history.

\paragraph{\texttt{lcm\_grep(pattern, summary\_id?).}}
Performs regex search across the full immutable message history. Returns matching messages grouped by the summary node that currently covers them, allowing the model to understand which region of the conversation each match belongs to. Results are paginated to prevent context flooding. An optional \texttt{summary\_id} parameter restricts the search to messages within the scope of a particular summary. This tool is available to both the main agent and sub-agents.

\paragraph{\texttt{lcm\_describe(id).}}
Returns metadata for any LCM identifier (file or summary). For file IDs, this includes the original path, MIME type, token count, and the exploration summary generated when the file was first encountered (Section~2.2). For summary IDs, this includes the summary kind (leaf or condensed), token count, parent pointers, and the full summary text. This tool allows the model to inspect what a compacted reference represents without expanding it.

\paragraph{\texttt{lcm\_expand(summary\_id).}}
Expands a summary node into its constituent messages, reversing the compaction that created it. Because expansion can recover arbitrarily large volumes of earlier conversation, this tool is restricted to sub-agents spawned via the \texttt{Task} tool; the main agent cannot call it directly. This restriction prevents uncontrolled context growth in the primary interaction loop. When the main agent needs to inspect compacted history, it delegates the expansion to a sub-agent, which processes the expanded content in its own context window and returns only the relevant findings.

\subsection{Operator Tools}

These tools implement Operator-Level Recursion (Section~3.1), allowing the model to process unbounded datasets via a single tool call while the engine handles iteration, concurrency, and retries.

\paragraph{\texttt{llm\_map(input\_path, prompt, output\_schema, ...).}}
Processes each item in a JSONL input file by dispatching it as an independent LLM API call. The engine manages a worker pool (default concurrency 16), validates each response against the caller-supplied JSON Schema, and retries failed items with feedback from the validation error. No tools or side effects are available to the per-item calls; each is a pure function from input to structured output. Results are written to a JSONL output file and registered in the immutable store. This tool is appropriate for high-throughput, side-effect-free tasks such as classification, entity extraction, or scoring.

\paragraph{\texttt{agentic\_map(input\_path, prompt, output\_schema, read\_only, ...).}}
Similar to \texttt{llm\_map}, but spawns a full sub-agent session for each item rather than a single LLM call. Each sub-agent has access to tools (file reads, web fetches, code execution) and can perform multi-step reasoning. A \texttt{read\_only} flag controls whether sub-agents may modify the filesystem. Output validation and retry logic follow the same pattern as \texttt{llm\_map}. This tool is appropriate when per-item processing requires tool use or multi-turn reasoning that cannot be captured in a single prompt.

\subsection{Delegation Tools}

These tools manage the spawning and coordination of sub-agents, enabling hierarchical task decomposition while preserving context isolation.

\paragraph{\texttt{Task(prompt, subagent\_type, delegated\_scope, kept\_work, ...).}}
Spawns a single sub-agent to execute a task autonomously. The sub-agent receives its own context window and session; all intermediate tool calls remain private to the sub-agent, and only its final answer is returned to the parent. This preserves the parent's context for orchestration rather than consuming it with intermediate results.

The \texttt{Task} tool inherits from OpenCode's sub-agent mechanism but introduces a key modification for LCM: an \textbf{infinite-recursion guard}. When a sub-agent (as opposed to the root agent) invokes \texttt{Task}, it must provide two additional parameters: \texttt{delegated\_scope}, describing the specific slice of work being handed off, and \texttt{kept\_work}, describing the work the caller retains for itself. If the caller cannot articulate what it is keeping (that is, if it would delegate its entire responsibility), the call is rejected with the instruction to perform the work directly. This forces each level of delegation to represent a strict reduction in scope, creating a natural bottoming-out condition analogous to a well-founded recursion. Read-only exploration agents are exempted from this check, as they lack the ability to spawn further sub-agents and thus cannot recurse.

\paragraph{\texttt{Tasks(tasks[]).}}
Accepts an array of two or more independent task descriptions and executes them as parallel sub-agents. \texttt{Tasks} dispatches all sub-agents concurrently and aggregates their results, preserving the parent's context from intermediate clutter. The recursion guard is not applied to parallel tasks, as they represent sibling decomposition (splitting work into independent units) rather than nested delegation. This tool was added in Volt to encourage parallel work decomposition as a first-class pattern, complementing the sequential delegation of \texttt{Task}.

\end{document}